\pdfoutput=1
\documentclass[11pt]{article}

\usepackage[]{acl}

\usepackage{times}
\usepackage{latexsym}

\usepackage[T1]{fontenc}
\usepackage[utf8]{inputenc}

\usepackage{microtype}
\usepackage{amssymb}
\usepackage{amsopn}
\usepackage{graphicx}
\usepackage{multirow}
\usepackage{makecell}
\usepackage[english]{babel}
\usepackage{bm}
\usepackage{setspace}
\usepackage{color,xcolor}

\usepackage{pifont}
\usepackage{bm}
\usepackage{amsfonts}

\usepackage{graphicx}
\usepackage[normalem]{ulem}
\usepackage{multirow}
\usepackage{amsmath}
\usepackage{url}

\usepackage{tikz}
\usepackage{xcolor}
\usepackage{tcolorbox}
\usepackage{color,xcolor}

\usepackage{amssymb}
\usepackage{amsopn}
\usepackage{graphicx}
\usepackage{multirow}
\usepackage[english]{babel}
\usepackage{bm}
\usepackage{array}
\usepackage{setspace}
\usepackage{todonotes}

\usepackage{xspace}
\usepackage{amsfonts}
\usepackage{array,multirow}
\usepackage{pgfplots}
\usepackage{tikz}
\usepackage{verbatim}
\usepackage{subfig}
\usepackage{arydshln}
\usepackage{booktabs}
\definecolor{ugreen}{rgb}{0,0.5,0}
\definecolor{mygreen}{RGB}{58,127,88}
\definecolor{iyellow}{RGB}{255,250,205}
\definecolor{ipurple}{RGB}{230,230,250}
\definecolor{myred}{RGB}{160,52,52} %238，44，44
\definecolor{myblue}{RGB}{30,144,255}
\definecolor{myorange}{RGB}{255,127,80}
\definecolor{mypurple}{RGB}{255,20,147}

\title{Task-guided Disentangled Tuning for Pretrained Language Models}

\author{Jiali Zeng$^{1}\thanks{\ \ Corresponding author.}$, \ Yufan Jiang$^{1}$, \ Shuangzhi Wu$^{1}$, \  Yongjing Yin$^{2}$, \ Mu Li$^{1}$ \\
$^{1}$Tencent Cloud Xiaowei, Beijing, China \\
$^{2}$Zhejiang University, Westlake University, Zhejiang, China\\
{\tt \{lemonzeng,garyyfjiang,frostwu,ethanlli\}@tencent.com} \\
{\tt yinyongjing@westlake.edu.cn} \\ 
}

\begin{document}
% \begin{CJK*}{UTF8}{gbsn}
\maketitle
\begin{abstract}

Pretrained language models (PLMs) trained on large-scale unlabeled corpus are typically fine-tuned on task-specific downstream datasets, which have produced state-of-the-art results on various NLP tasks.
However, the data discrepancy issue in domain and scale makes fine-tuning fail to efficiently capture task-specific patterns, especially in the low data regime.
To address this issue, we propose {\bf T}ask-guided {\bf D}isentangled {\bf T}uning ({\bf TDT}) for PLMs, which enhances the generalization of representations by disentangling task-relevant signals from the entangled representations. 
For a given task, we introduce a learnable confidence model to detect indicative guidance from context, and further propose a disentangled regularization to mitigate the over-reliance problem.
% disentangled regularization to encourage the model to learn generalized representations to mitigate the over-reliance problem.
% We apply TDT to different PLMs and evaluate on various downstream tasks. 
Experimental results on GLUE and CLUE benchmarks show that TDT gives consistently better results than fine-tuning with different PLMs, and extensive analysis demonstrates the effectiveness and robustness of our method.
Code is available at https://github.com/lemon0830/TDT.
% We will release code upon acceptance.

% Fine-tuning has become the de facto way to use pretrained language models (PLMs) to perform downstream tasks.
% Typically, PLMs are trained on large-scale unlabeled corpus and finetuned on limited task-specific dataset. 
% The data discrepancy in domain and scale make fine-tuned models fail to capture task-specific patterns.
% In this paper, we propose {\bf T}ask-guided {\bf D}isentangled {\bf T}uning ({\bf TDT}) for PLMs. 
% For a given task, we introduce a learnable confidence model, which aims to detect indicative guidance from the context and helps the PTM better understand downstream tasks.
% Meanwhile, we further enhance our method with a 
% disentangled regularization to encourage the model to learn generalized representations to mitigate the over-reliance problem.
% We apply TDT to different PLMs and evaluate on various downstream tasks. Experimental results and extreme analysis demonstrate the effectiveness and robustness of our proposed TDT. 
\end{abstract}

\section{Introduction}
% rewrite

% Recently, pretrained language models (PLMs) such as BERT\cite{devlin-etal-2019-bert}, Roberta \cite{liu2019roberta} and ALBERT \cite{lan2019albert}
% GPT-3?
% have enabled many breakthroughs in natural language processing. 
% have proved effective in a variety of language understanding tasks benefiting form large-scale unlabeled documents.
% Commonly, PLMs first learn general linguistic and semantic knowledge from massive general corpus like Wikipedia and Book corpus, which are then adapted to downstream tasks via supervised fine-tuning.
% Generally, a task-specific head on top of PLMs are added and updated with the PLMs parameters during fine-tuning stage. 

In recent years, pretrained language models (PLMs) trained in a self-supervised manner like mask language modeling have achieved promising results on various natural language processing (NLP) tasks \cite{devlin-etal-2019-bert,yang2019xlnet,liu2019roberta}, which learn general linguistic and semantic knowledge from massive general corpus.
To adapt PLMs to specific NLP tasks, a commonly-used approach is fine-tuning, where the whole or part of model parameters are tuned by task-specific objectives.
% nd then adapt themselves to the downstream tasks via supervised fine-tuning. %Generally, a task-specific head on top of PLMs are added and updated with the PLMs parameters during fine-tuning stage.
Despite its success, the fine-tuned models have been proven ineffective to capture task-specific patterns due to the gap between task-agnostic pre-training and the weak fine-tuning with limited task-specific data \cite{gu-etal-2020-train, gururangan-etal-2020-dont, kang-etal-2020-neural}.

\begin{figure}[!th]
\centering
\includegraphics[width=1.0\linewidth]{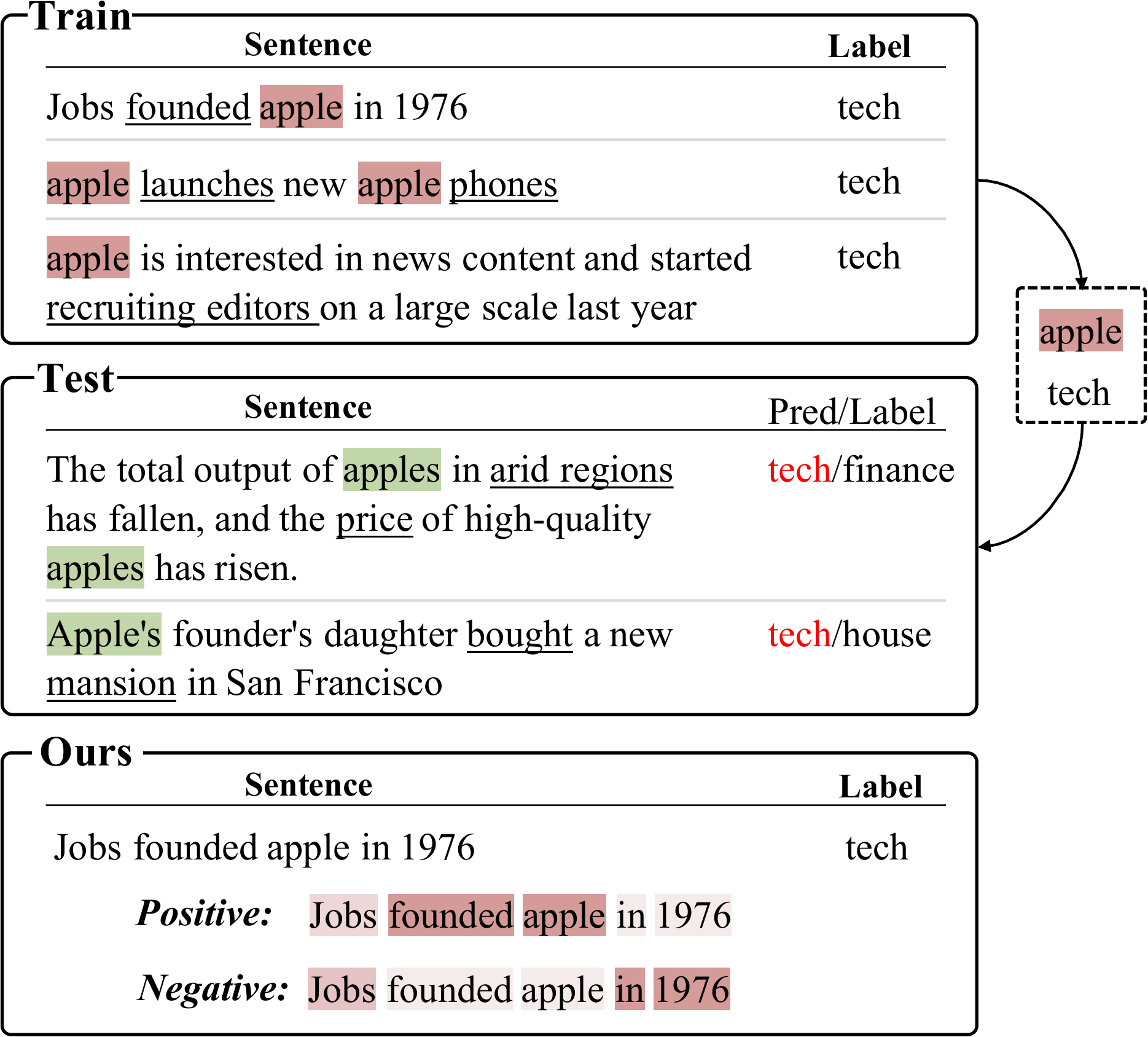}
\caption{
{\bf An over-reliance example of news classification task}. The fine-tuned models tend to learn a simple rule that ``Apple'' (red) indicates ``tech'' class while ignore the real meaning of ``apples'' (green).
A reliable model is expected to find out truly task-specific patterns (underlined words) instead of some high frequency but insignificant words (``apple''). 
}
\label{fig_example}
\end{figure}

% Despite its great success, the fine-tuned models are proved inefficient to capture task-specific patterns due to the gap between task-agnostic pre-training and the weak fine-tuning with limited labeled data \cite{gu-etal-2020-train, gururangan-etal-2020-dont, kang-etal-2020-neural}.

To address this problem, most existing methods focus on adapting PLMs to downstream tasks by continual pre-training on in-domain unsupervised data \cite{gururangan-etal-2020-dont,gu-etal-2020-train, wu-etal-2021-domain,kang-etal-2020-neural,ye2021influence}. 
For example, \citet{gu-etal-2020-train} propose intermediate continual pre-training with a selective masking strategy, and \citet{gururangan-etal-2020-dont} adapt PLMs to in-domain tasks by domain-adaptive pretraining. 
Although straightforward, these kinds of methods heavily rely on the selection of large-scale additional domain corpora and the design of appropriate intermediate training tasks \cite{wang-etal-2019-tell,aghajanyan2021muppet}. 
% In addition, they necessitate collecting domain corpus for each downstream task, which is inefficient . 

In this paper, we propose a {\bf T}ask-guided {\bf D}isentangled {\bf T}uning ({\bf TDT}) for PLMs by automatically detecting task-specific informative inputs without the need of additional corpora and intermediate training. 
The core component of TDT is a confidence model which assigns each token a confidence score, and we construct distilled samples by retaining informative tokens with high confidence scores while perturbing the rest. 
The confidence model performs a ``deletion game'' strategy, which encourages the model to perturb inputs as much as possible and to maintain the performance of downstream tasks to the greatest extent with the distilled samples. 
% In this way, we are able to find informative inputs for a certain downstream task.
Although the informative tokens are important for downstream predictions, existing work shows that over-relying on part of these words may result in pool generalization, i.e., over-reliance problem \cite{moon2020masker,geirhos2020shortcut,sun2019shortcutnlp}. Take the sentences in Figure \ref{fig_example} as an example, when the context word ``Apple" frequently co-occurs with the label ``tech", fine-tuned models may learn a spurious association by binding ``Apple" and ``tech", leading to incorrect predictions of sentences which contain ``apple'' but belong to other categories. 

% Further, we enhance our method with a disentangled regularization, which .
% To help fine-tuned models learn more generalizable representations by disentangling task-relevant signals from task-irrelevant ones
% Further, we enhance our method with a disentangled regularization, which .
Based on the observation, we further enhance our method with a disentangled regularization, aiming to distinguish task-relevant and task-irrelevant features.
% enhances the generalization of  
% To detach task-relevant and task-irrelevant features, we further enhance our method with a disentangled regularization.
First, we construct two variants of the original input in a complementary view: (1) {\it positive} variant, which maintains the high-confidence keywords, and (2) {\it negative} variant, derived by a ``cut-out-keyword'' operation on the original input. 
Next, we propose a ``triplet-style loss'', which makes predictions between the {\it original} input and the {\it positive} variant similar while the predictions between the {\it negative} variant and the other two different. 
To illustrate the mechanism of our disentangled regularization, 
% back to Figure \ref{fig_example}, 
we go back to Figure \ref{fig_example} and take the sentence ``Jobs founded apple in 1976'' 
% from the news classification task 
as an example.
% We encourage the consistency between the {\it original} input and the {\it positive} variant, and maximize the difference between the {\it negative} variant and the other two. 
% We keep the consistency between the {\it original} input and the {\it positive} variant, while differentiating the {\it negative} variant and the other two as far as possible. 
% In this manner, 
Under the influence of the disentangled regularization, the {\it positive} variant tends to retain 
% complete and precise context 
clue words for predictions (i.e., ``founded apple''), while the {\it negative} variant, as the complement (i.e., ``Jobs in 1976''), tends to be task-irrelevant. 
% The disentangled regularization prompts the model to distinguish task-relevant context from task-irrelevant context, and improves the model's generalization.

We evaluate our TDT on a wide range of neural language understanding benchmark datasets in English and Chinese, i.e., GLUE and CLUE, and our TDT affords strong predictive performance compared with standard fine-tuning.
% across a spectrum of downstream tasks.
Moreover, we conduct extensive analysis with respect to robustness to perturbation, domain generalization, and low-resource settings, from which we conclude:
\begin{itemize}
    \item TDT learns reasonable confidence scores for input tokens.
    \item TDT is robust to input perturbation and domain shift by encouraging the model to learn more generalized features.
    \item TDT effectively captures the high-confidence decisive cues for downstream tasks, thus alleviating over-fitting in low-resource scenarios.
\end{itemize}
%Finally, we compared TDT with previous methods to demonstrate the effectiveness.
%Notably, our method is model-agnostic and can be integrate with other methods to potentially achieve further improvements.

\begin{figure*}[!h]
\centering
\includegraphics[width=0.85\linewidth]{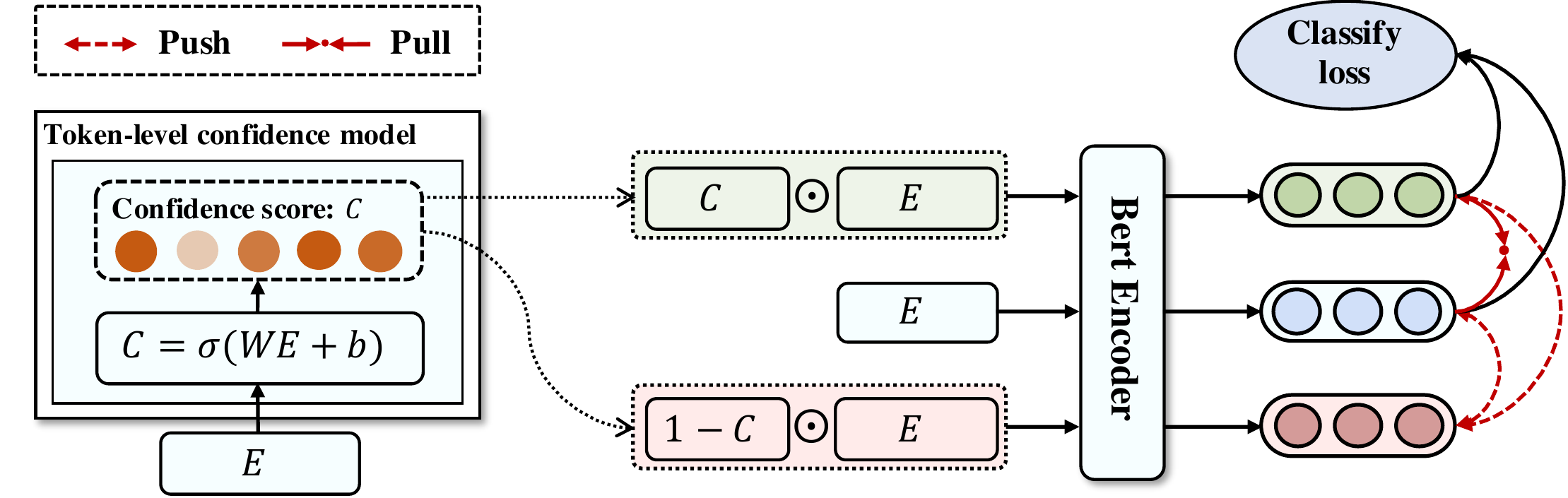}
\caption{
The overall framework of our proposed Task-guided Disentangled Tuning method.
}
\label{fig_model}
\end{figure*}

\section{Method}
In this section, we begin with a brief introduction of the vanilla Fine-tuning, and then introduce {\bf T}ask-guided {\bf D}isentangled {\bf T}uning (\textbf{TDT}) in detail.
Figure \ref{fig_model} shows the overall framework. 
\textbf{TDT} is composed of two parts: (1) token-level confidence model, which discovers the essential parts of inputs for the model prediction; (2) task-guided regularization, which promotes the model to decouple task-relevant keywords from non-keyword context.
% , which is shown in Figure . 

\subsection{Vanilla Fine-tuning}

Given an example of training data $<X,y>$, where $X$=$\{x_1, ..., x_i, ..., x_n\}$ is the input sequence and $y$ is its corresponding label. 
We first map each token $x_i$ to a real-valued vector $e_i$ by an embedding layer. 
Then, the packed embedding output $E$=\{$e_i$\} is fed into the PLM to get the contextualized sentence representations $H$=\{$h_{cls}, h_1, ..., h_n$\}, and the hidden state $h_{cls}$ is used to conduct classification with a MLP head.
We fine-tune the parameters of the PLM with the cross entropy loss:
\begin{equation}
    \mathcal{L}_{cla} = -{\rm log} P(y|H).
\end{equation}

\subsection{Token-level Confidence Model} \label{sec_confience}
% We aim to discover which inputs are essential for the model prediction and then encourage a stronger encoder to conduct effective fine-tuning for various tasks.
For each token $x_i$, we generate a scalar $c_i \in [0,1]$, coined confidence score, by stacking a single-layer feed-forward network with {\it sigmoid} activation on the top of the embedding layer:
\begin{equation}
    c_i = \sigma(We_i + b), \label{eq_confidence}
\end{equation}
where $W$ and $b$ are trainable parameters.
Based on the confidence score, we obtain a distilled sample \{$e^+_i$\} defined as
\begin{equation}
    e^+_i = c_i \odot e_i + (1-c_i) \odot \mu_{0}, \label{eq_positive}
\end{equation}
where $\mu_{0}$ is a perturbation term and $\odot$ denotes element-wise multiplication. 
Specifically, the perturbation term $\mu_{0}$ can be a zero vector, a random Gaussian noise vector, or the average of the token embedding, and we choose the last one in this paper.
In this manner, for the distilled sample of each training instance,
the higher the $c_i$ is, the more semantic information of the $i$-th token retains, while the tokens with lower scores are perturbed.
% the token $x_i$ retrains more semantic information with higher $c_i$, 

Then, the distilled sample \{$e^+_i$\} is fed into the PLM to generate the sentence representations $H^+$=$\{h^+_i\}$.
Inspired by ``deletion game'' \cite{Fong2017interpretable, voita-etal-2019-analyzing},
% given the representations $H^+$ and the confidence score $c_i$,
the objective function of the confidence model is
\begin{equation}
    \mathcal{L}_C = -{\rm log} P(y|H^+) + \gamma ||C||_2,
\end{equation}
where $C=\{c_i\}$ is the set of confidence scores of $X$.
The first term is the cross entropy loss of classification on the distilled sample to encourage the confidence model to assign higher scores to the more decisive part of the input, and the second term 
serves as a penalty to 
prevent the model from mode collapsing (i.e., always choosing $c_i$=$1$).
% Here, we simply set the $\gamma$ as 1e-4. (put it on the experiment)

\subsection{Task-Guided Regularization} \label{sec_regularization}
It has been widely observed that the pretrained models tend to learn an easy-to-learn but not generalizable solution by vanilla fine-tuning on various NLP tasks \cite{sun2019shortcutnlp, mccoy-etal-2019-right,min-etal-2019-compositional,niven-kao-2019-probing}.
% To learn a reliable encoder to capture generalize features and then benefit the classification,
To alleviate this issue, we further propose a triplet-style loss on the model predictions.

Specifically, for each input sequence, we derive two different variants: a {\it positive} variant and a {\it negative} variant.
The {\it positive} variant is expected to maintain the most informative tokens to task prediction
%, while the negative one is verse versa.
and vice versa.
As aforementioned, our confidence model removes the meaningless tokens
% to ameliorate the effective fine-tuning\
by setting the corresponding confidence scores to zero.
% In other words, a smaller score means a larger perturbation, namely a less significant impact on the prediction.
Based on the confidence scores, we directly treat the distilled sample generated by Eq. \ref{eq_positive} as the {\it positive} variant and generate the {\it negative} variant as
\begin{equation}
    e^-_i = (1-c_i) \odot e_i.
\end{equation}
Given the original input and the two derived variants, we feed them into the PLM with the classifier, and obtain three prediction distributions $P(y|H)$, $P(y|H^+)$, and $P(y|H^-)$.
Finally, we regularize these distributions by a triplet ranking loss
\begin{align}
    \nonumber \mathcal{L}_R = \max(m + & d(P(y|H^+), P(y|H)) \ - \\ 
    \nonumber & d(P(y|H^-), P(y|H)) \ - \\ 
    & d(P(y|H^-), P(y|H^+)), 0)
\end{align}
where $m$ is a hyperparameter indicating a margin for the loss and $d(\cdot)$ denotes the Kullback-Leibler (KL) divergence.
By minimizing $\mathcal{L}_R$, the positive variant will be closer to the original input while the negative variant will be farther from the other two.
% We minimize $\mathcal{L}_R$, which pushes $d(P(y|H^+)$,$P(y|H))$ to 0, and pull $d(P(y|H^-)$,$P(y|H))$ and $d(P(y|H^-)$, $P(y|H^+))$ to be greater than $d(P(y|H^+)$, $P(y|H))$+margin.
Thus, the model is encouraged to disentangle task-relevant signals from task-irrelevant factors, and generate more general representations.

\begin{table*}[!ht]
\centering
\small
\begin{spacing}{1.2}
\begin{tabular}{lccccccccc}
\toprule
{\bf Model} & {\bf MNLI} & {\bf QQP} & {\bf QNLI} & {\bf SST-2} & {\bf CoLA} & {\bf STS-B} & {\bf MRPC} & {\bf RTE} & {\bf Avg} \\
\midrule
\multicolumn{10}{l}{\it BERT-base} \\
% FineTuning \dag & 83.7 & 90.8 & 89.3 & 91.7 & 48.9 & 91.0 & 86.3 & 71.4 & 81.64\\
FineTuning & 84.5 & 90.9 & 91.3 & 92.8 & 60.5 & 88.7 & 85.1 & 67.5 & 82.66 \\
{\bf TDT} & 85.3 & 91.2 & 91.9 & 93.7 & 62.4 & 89.3 & 87.5 & 71.8 & \bf{84.14} \\
\midrule
\multicolumn{10}{l}{\it BERT-large} \\
FineTuning \dag & 86.6 & 91.3 & 92.3 & 93.2 & 60.6 & 90.0 & 88.0 & 70.4 & 84.05 \\
FineTuning & 85.9 & 90.9 & 92.3 & 93.9 & 61.5 & 90.0 & 86.0 & 75.1 & 84.45 \\
{\bf TDT} & 86.4 & 91.4 & 92.6 & 94.3 & 66.2 & 89.9 & 88.5 & 75.8 & \bf{85.64} \\
\midrule
\multicolumn{10}{l}{\it RoBERTa-large} \\
FineTuning \dag & 90.2 & 92.2 & 94.7 & 96.4 & 68.0 & 92.4 & 90.9 & 86.6 & 88.92 \\
FineTuning & 90.5 & 92.3 & 94.4 & 96.6 & 67.4 & 92.2 & 91.9 & 87.7 & 89.13 \\
{\bf TDT} & 90.6 & 91.9 & 94.7 & 97.0 & 69.3 & 92.5 & 93.1 & 91.0 & \bf{90.01} \\
\midrule
XLNet \dag  & 90.8 & 92.3 & 94.9 & 97.0 & 69.0 & 92.5 & 90.8 & 85.9 & 89.15 \\
ELECRTA \dag & 90.9 & 92.4 & 95.0 & 96.9 & 69.1 & 92.6 & 90.8 & 88.0 & 89.46 \\
DeBERTa \dag & 91.1 & 92.4 & 95.3 & 96.8 & 70.5 & 92.6 & 91.9 & 88.3 & 89.86 \\
ALBERT \dag & 90.8 & 92.2 & 95.3 & 96.9 & 71.4 & 93.0 & 90.9 & 89.2 & 89.96 \\
\bottomrule 
\end{tabular}
\end{spacing}
\caption{
\label{tab_results_glue}
{\bf Experimental results on GLUE language understanding benchmark}.
When take RoBERTA-large as the PLM, for RTE and STS, we follow \citet{liu2019roberta} to finetune starting from the MNLI model instead of the baseline pretrained model.
Methods with {\dag} denote that we directly report the scores from corresponding paper, and others are from our implementation.
}
\end{table*}

\subsection{Overall Training Objective}
The final training objective is
\begin{equation}
    \mathcal{L} = \mathcal{L}_{cla} + \alpha \mathcal{L}_C + \beta \mathcal{L_R},
\end{equation}
where $\alpha$ and $\beta$ are non-negative hyper-parameters 
% weights assigned beforehand 
to balance the effect of each loss term.

\section{Experiments}

\begin{table*}[!ht]
\centering
\small
\begin{spacing}{1.2}
\begin{tabular}{lcccccc}
\toprule
\multirow{2}{*}{\bf Task} & \multicolumn{2}{c}{\it BERT-wwm-base} & \multicolumn{2}{c}{\it MacBERT-large} & \multicolumn{2}{c}{\it RoBERTa-wwm-large} \\
\cmidrule(r){2-3} \cmidrule(r){4-5} \cmidrule(r){6-7}
& FineTuning & TDT & FineTuning & TDT & FineTuning & TDT \\
\midrule
{\bf OCNLI} & 74.6 & {75.3} & 78.3 & {79.8} & 78.1 & {79.5} \\
{\bf IFLYTEK} & 60.8 & {62.2} & 61.5 & {61.8} & 61.8 & {62.9} \\
{\bf CSL} & 84.7 & {85.5} & 86.8 & {87.0} & 86.1 & {87.2} \\
{\bf TNEWS} & 56.9 & {57.3} & 58.5 & 58.7 & 59.0 & {59.2} \\
{\bf AFQMC} & 74.0 & {75.0} & 76.2 & {76.8} & 76.0 & {76.2} \\
\midrule
{\bf Avg} & 70.20 & \bf{71.06} & 72.26 & \bf{72.82} & 72.20 & \bf{73.00} \\ 
\bottomrule
\end{tabular}
\end{spacing}
\caption{
\label{tab_results_clue}
{\bf Experimental results on CLUE language understanding benchmark}.
For TNEWS, we only use the raw ``sentence'' for classification without the ``keywords'' information. 
For CSL, we only mask the ``abst'' sequence and keep the ``keywords'' sequence unchanged in our proposed method.
}
\end{table*}

\subsection{Datasets}
We evaluate our proposed method by fine-tuning the pretrained models on the General Language Understanding Evaluation (GLUE) \cite{wang-etal-2018-glue} and the Chinese Language Understanding Evaluation (CLUE) \cite{xu2020CLUE}.
Concretely, the GLUE benchmark has 8 different text classification or regression tasks including MNLI, MRPC, QNLI, QQP, RTE, SST-2, SST-B, and CoLA.
The CLUE benchmark includes 9 tasks spanning several single-sentence/sentence-pair classification tasks, and we choose 5 tasks, OCNLI, IFLYTEK, CSL, TNEWS, and AFQMC.
% the short text classification task TNEWS, the long text classification tasks IFLYTEK and CSL, and sentence-pair classification tasks AFQMC and OCNLI.
The detailed data statistics and metrics are provided in Appendix \ref{sec:appendixA}.

\subsection{Model \& Training}
We use the pretrained models and codes provided by HuggingFace\footnote{https://github.com/huggingface/transformers}. 
We take BERT-base \cite{devlin-etal-2019-bert}, BERT-large \cite{devlin-etal-2019-bert} and RoBERTa-large \cite{liu2019roberta} as our backbones on GLUE, while BERT-wwm-base \cite{cui2019pre}, MacBERT-large \cite{cui-etal-2020-revisiting}, and RoBERTa-wwm-large \cite{cui2019pre} on CLUE.
We tune the task specific hyper-parameters $m \in \{0, 2\}$, $\alpha \in \{0.5, 2, 4\}$ and $\beta \in \{0.5, 1\}$.
Detailed experimental setups are shown in Appendix \ref{sec:appendixB}.
Following previous work \cite{Lee2020Mixout,aghajanyan2020better}, we report results of the development sets, since the performance on the test sets is only accessible on the leaderboard with a limitation of the number of submissions.

\subsection{Main Results}
\paragraph{Results on GLUE.}
We illustrate the experimental results on the GLUE benchmark in Table \ref{tab_results_glue}.
We can observe that the PLMs enhanced by \textit{TDT} outperforms \textit{FineTuning} by a large margin across all the tasks.
Specifically, \textit{TDT}s achieve 1.48 points, 1.19 points and 0.88 points (on average) improvement over BERT-base, BERT-large, and RoBERTa-large, respectively.
In particular, {\it BERT-base+TDT} achieves competitive performance compared with {\it BERT-large+FineTuning}, showing that our method is more efficient to find task-specific information for downstream tasks. 
% This may be because our training strategy prompts the model to conduct prediction with as much as less information, while isolating the representations of the task-relevant signal from task-irrelevant factors.
This may be because our training strategy prompts the models to predict with as little information as possible, isolating the task-related signals from the whole representations.

{\it RoBERT-large} trained with \textit{TDT} surpasses XLNet-large \cite{yang2019xlnet} ALBERT-xxlarge \cite{lan2019albert}, DeBERTa-large \cite{he2020deberta}, and ELECTRA-large \cite{clark2020electra}, which are specially designed with different architectures and pre-training objectives.

\begin{figure}[!t]
\centering
\includegraphics[width=0.95\linewidth]{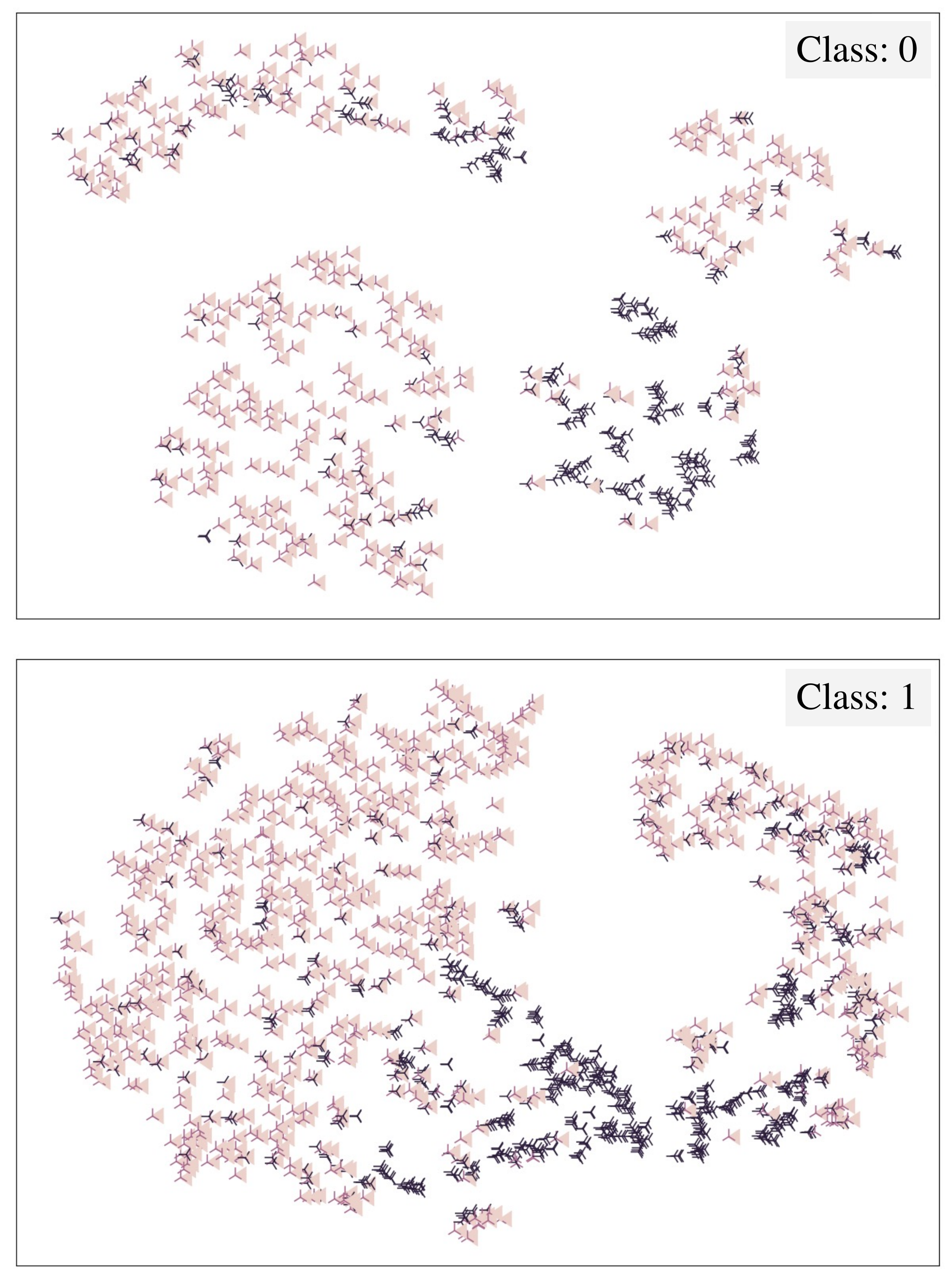}
\caption{
{\bf Visualization of representations} of original input and two derived variants, where the triangle-shaped (pink), tri-up-shaped (purple), and tri-left-shaped (black) points denote the representations of {\color{pink}\emph{original input}}, {\color{purple}\emph{positive variants}}, and {\color{black}\emph{negative variants}}, respectively.
}
\label{fig_analy_visual}
\end{figure}

\paragraph{Results on CLUE.}
Table \ref{tab_results_clue} shows the overall results on the 5 tasks of CLUE benchmark. 
% Compared with \textit{FineTuning}, our \textit{TDT} consistently yields better results on all tasks.
Concretely, \textit{TDT} significantly outperforms \textit{FineTuning} on {CSL}, {IFLYTEK}, {AFQMC}, and {OCNLI}, and shows competitive results on the short text classification task {TNEWS}, indicating the advantage of extracting important parts from long text or multiple input sequences.
Note that TNEWS generally requires additional knowledge (e.g., keywords) as a supplement due to the short input, and thus cannot show the superiority of \textit{TDT}.
% The underlying reason is that our \textit{TDT} is good at extracting important parts from long text or multiple input sequences.
% our method cannot show advantages in such a scenario.

\section{Analysis \& Discussion}

% In order to better understand ABC, we conduct empirical analysis and further launch a discussion on the difference of our method and prior methods.

\subsection{Visualization of Representations}
In Figure \ref{fig_analy_visual}, we plot t-SNE visualizations \cite{maaten2008sne} of three kinds of representations generated by BERT-large trained with TDT on CoLA dev set.
% We can see that the original input representations and the positive variant representations in the same class are close to each other.
We can see that the representations of the original input are close to those of the positive variant in the same class.
% are close to each other.
Although the negative variant representations are really similar to the original ones which derive the former, they are clearly separated from the other representations.
The learned disentangled representations reveal that {\bf the model trained with TDT is able to distinguish task-specific keywords and non-keyword context, which plays an important role in increasing models' robustness}.

\begin{figure}[!t]
\centering
\includegraphics[width=0.85\linewidth]{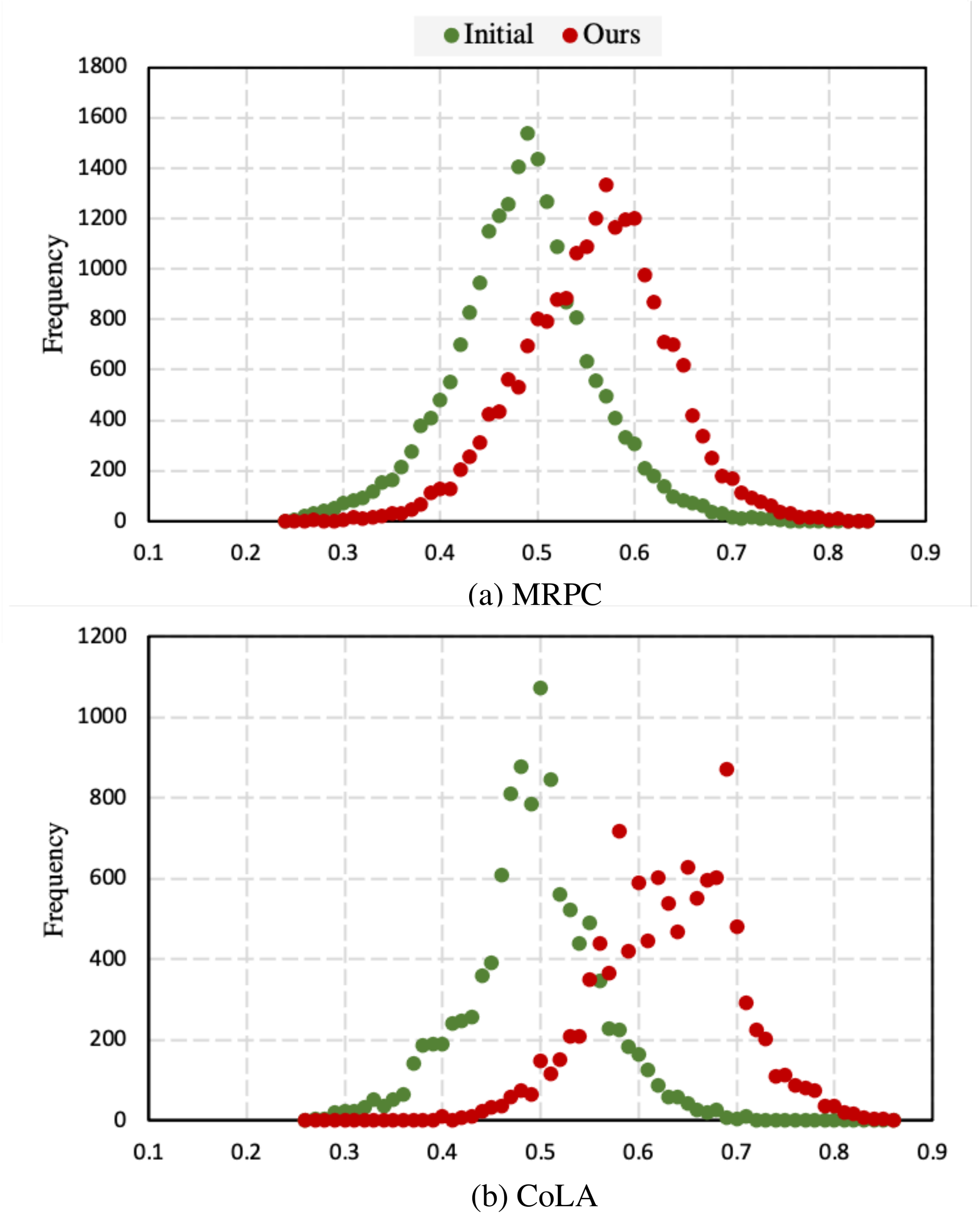}
\caption{
{\bf Distribution of confidence scores} on MRPC and CoLA dev sets.
}
\label{fig_analy_distribution}
\end{figure}

\begin{figure*}[!t]
\centering
\includegraphics[width=1.0\linewidth]{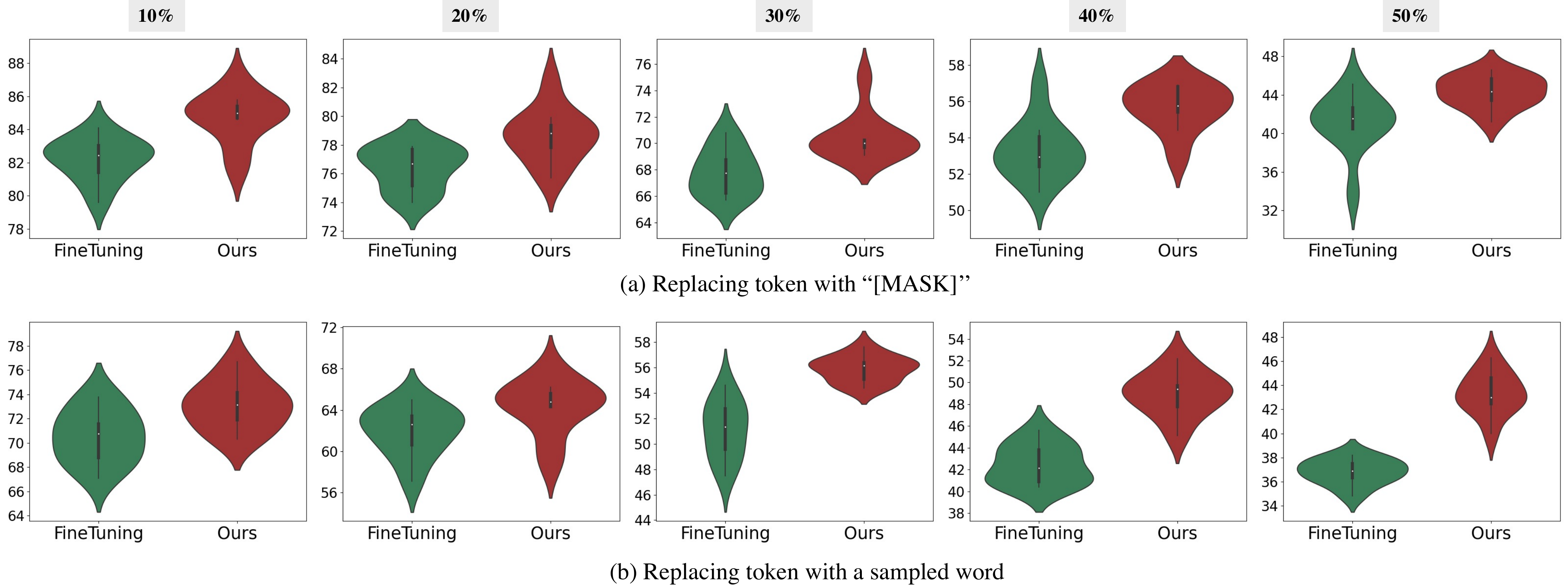}
\caption{
{\bf Robustness to Input Perturbation}.
The Y-axis is the accuracy on the development set.
% Across 10 random seeds of our runs were higher using our method than standard fine-tuning.
}
\label{fig_analy_robustness}
\end{figure*}

\begin{figure}[!t]
  \centering
  \vspace{2pt}
  \begin{tikzpicture}
    \footnotesize{
    \begin{axis}[
    ymajorgrids,
    xmajorgrids,
    grid style=dashed,
    width=.47\textwidth,
    height=.35\textwidth,
    legend style={at={(0.43,0.85)}, anchor=south west},
    xlabel={\footnotesize{dropout rate \%}},
    ylabel={\footnotesize{Accuracy}},
    ylabel style={yshift=-1em},xlabel style={yshift=0.0em},
    yticklabel style={/pgf/number format/precision=1,/pgf/number format/fixed zerofill},
    ymin=53,ymax=93, ytick={60, 70, 80, 90},
    xmin=5,xmax=65,xtick={10,20,30,40,50, 60},
    legend style={yshift=-6pt, legend columns=2,legend plot pos=right,font=\scriptsize,cells={anchor=west},at={(0.05,0.15)}}
    ]

    \addplot[mygreen!100,mark=diamond*,line width=1pt] coordinates {(10,80.6) (20,80.6) (30,76.2) (40,67.6)  (50,61.5) (60,55.9)};
    \addlegendentry{\scriptsize FineT}
    \addplot[mygreen!100,mark=diamond*,line width=1pt, densely dotted] coordinates {(10,84.5) (20,81.6) (30,81.4) (40,75.2)  (50,65.9) (60,56.1)};
    \addlegendentry{\scriptsize FineT-R}
    \addplot[myred!100,mark=otimes*,line width=1pt] coordinates {(10,82.1) (20,82.6) (30,79.7) (40,69.1)  (50,65.2) (60,58.8)};
    \addlegendentry{\scriptsize Ours}
    \addplot[myred!100,mark=otimes*,line width=1pt, densely dotted] coordinates {(10,87) (20,83.3) (30,82.8) (40,78.4)  (50,69.8) (60,63)};
    \addlegendentry{\scriptsize Ours-R}

    \end{axis}
    }
  \end{tikzpicture}
  \caption{
  {\bf Accuracy of BERT-large trained with different methods and evaluated on MPRC dev set with different drop rates}.
%   Due to the limited space, 
  We denote vanilla fine-tuning as FineT.
  The solid lines indicate results on the datasets constructed by dropping tokens in descending order of confidence scores.
  The dotted lines denotes results on the datasets constructed by dropping tokens in increasing order of confidence scores.
}\label{fig_analy_drop_by_mask}
\end{figure}
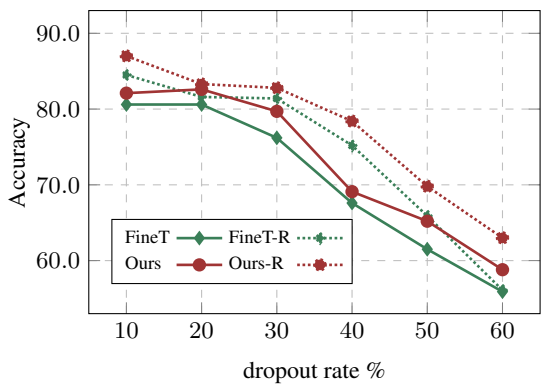

\subsection{Distribution of Confidence Scores}

We investigate the learned confidence score distributions in Figure \ref{fig_analy_distribution}.
It shows that although the initial distribution is consistent, the model learns different task-specific patterns (confidence distributions) on different tasks.

\subsection{Does our Confidence Model make a meaningful estimation for input tokens?} \label{sec_analy_drop_by_mask}

In section \ref{sec_confience}, we mention that TDT uses a scalar for evaluating the contribution of each input token.
To analyze whether the strategy can successfully learn a meaningful importance estimation,
we construct two sets of datasets based on MRPC dev set and then evaluate the performance of BERT-large with TDT and standard fine-tuning.
Specifically, we convert the confidence scores to probability distributions.
We generate the first set of datasets by dropping input tokens in descending order of the distributions  and generate the second set in ascending order.
In order to ensure language fluency, we replace each dropped token with a ``[MASK]'' token.
The results are shown in Figure \ref{fig_analy_drop_by_mask} and we observe that:
\begin{itemize}
    \item \textbf{TDT is more robust to incomplete input compared with Fine-tuning.}
    Specifically, although the performance of both {\it FineTuning} and {\it TDT} drops with the increase of dropout rate, our {\it TDT} achieves significantly better performance than {\it FineTuning} over all datasets.
    
    \item \textbf{Our learned confidence scores make reasonable assessments for each input token.}
    Particularly, regardless of the dropout rates and the training methods, dropping input tokens by the descending order of the masking scores always leads to worse performance. 
    % It shows that our learned masking score is reasonable 
\end{itemize}

\subsection{Robustness to Input Perturbation}
Based on the observation in Section \ref{sec_analy_drop_by_mask}, we further investigate the robustness of TDT on perturbed data.
To construct perturbed data, we use the dev set of MRPC and possibly replace the input at each position with a ``[MASK]'' token or a token sampled from the input sequence.
For each dropout rate, we construct 10 datasets with different random seeds and draw violin plots of the performance of BERT-large trained with TDT and fine-tuning (Figure \ref{fig_analy_robustness}).
We can see that {\it Ours} is consistently better than {\it FineTuning} in all groups, indicating the superior robustness to noisy data.
% The reason is that \textbf{our method encourages models to filter task-irrelevant information during training and thus the models are more robust to noisy data}.

\begin{table}[!t]
\centering
\small
\begin{spacing}{1.2}
\begin{tabular}{lccc}
\toprule
{\bf Task} & FineTuning & TDT & \textbf{$\Delta$} \\
\midrule
\multicolumn{4}{c}{\textbf{MNLI} (\textit{BERT-large})} \\
{MNLI-m} & 85.8 & 86.4 & \textbf{+0.6}  \\
{QQP} & 73.1 & 74.2 & \textbf{+1.1} \\
\midrule
\multicolumn{4}{c}{\textbf{OCNLI} (\textit{MacBERT-large})} \\
{CMNLI} & 70.6 & 71.8 & \textbf{+1.2}  \\
{BUSTM} & 64.8 & 66.4 & \textbf{+1.6} \\
\bottomrule
\end{tabular}
\end{spacing}
\caption{
\label{tab_results_generalze}
{\bf Performance of Domain Generalization}. The models are trained on MNLI/OCNLI but tested on out-of-domain data.
}
\end{table}

\subsection{Domain Generalization}
We evaluate how well the trained models generalizes to out-of-domain data on MNLI and OCNLI, Natural Language Inference (NLI) tasks of GLUE and CLUE respectively.
In detail, we fine-tune BERT-large on MNLI, and test the accuracy of the fine-tuned models on other NLI datasets in different domains including MNLI-mismatch\footnote{MNLI-mismatch has different domains from MNLI training data} and QQP.
% \footnote{QQP has two labels, \textit{duplicate} and \textit{not duplicate}. We map \textit{entailment} to \textit{duplicate} and map both \textit{neutral} and \textit{contradiction} to \textit{not duplicate}}.
Besides, we fine-tune MacBERT-large on OCNLI and conduct an evaluation on CMNLI\footnote{An NLI task of CLUE.} 
and BUSTM\footnote{A short text matching task of FewCLUE \cite{Xu2021Fewclue}}.
Detailed of Label Mapping is provided in Appendix \ref{sec:appendixC}.
% We map \textit{entailment} to \textit{0}, and map both \textit{neutral} and \textit{contradiction} to \textit{1}.}.
As Tabel \ref{tab_results_generalze} illustrates, TDT outperforms vanilla fine-tuning across different out-of-domain datasets.
The results suggest that \textbf{TDT encourages the model to learn more generalized features rather than some superficial contextual cues unique to training data}.

\begin{table}[!t]
\centering
\small
\begin{spacing}{1.2}
\begin{tabular}{lccc}
\toprule
{\bf Task} & FineTuning & TDT & \textbf{$\Delta$} \\
\midrule
\multicolumn{4}{c}{\textbf{CLUE} (\textit{MacBERT-large})} \\
OCNLI & 60.85 ($\pm$2.66) & 63.38 ($\pm$0.90) & \textbf{+2.53} \\
IFLYTEK & 54.12 ($\pm$0.75) & 54.78 ($\pm$0.94) & \textbf{+0.66} \\
CSL & 80.25 ($\pm$1.36) & 81.45 ($\pm$0.62) & \textbf{+1.20} \\
TNEWS & 53.50 ($\pm$0.58) & 53.33 ($\pm$0.25) & \textbf{-0.17} \\
AFQMC & 64.77 ($\pm$3.87) & 66.45 ($\pm$0.93) & \textbf{+1.68} \\
\midrule
\textbf{Avg} & 62.70 & 63.88 & \textbf{+1.18} \\
\bottomrule
\end{tabular}
\end{spacing}
\caption{
\label{tab_results_low_resource}
{\bf Experimental results in low-resource scenarios}.
We run 4 times for each task with different random seeds and report the average accuracy and the standard deviation.
}
\end{table}

\subsection{Results in Low-resource Scenarios}
Fine-tuning PLMs on very small amount of training data can be challenging and result in unstable performance due to the serious over-fitting issue.
In this section, we explore the effectiveness of TDT in such scenarios.
For each dataset in CLUE, we use MacBERT-large and sample 1k training examples as its training data.
As Table \ref{tab_results_low_resource} demonstrates, {\it TDT} improves the accuracy by 1.18 on average and reduces the standard deviation by up to 2.94. 
It suggests that \textbf{our TDT is more stable and efficient than vanilla fine-tuning when training PLMs on limited data}.
%  for {\it FineTuning}, the s of accuracy are large in 4 runs.
% Compared with {\it FineTuning}, {\it TDT} and alleviate unstable.
% It suggests that \textbf{although over-fitting is quite severe in low-resource scenarios, TDT can still effectively improve the model performance}.

\subsection{Compared with Variants}
\paragraph{Ablation Studies.}
We first conduct ablation studies to explore the effectiveness of two additional loss functions introduced in this paper and show the results in Table \ref{tab_results_discuss}.
We find that removing any of them leads to a performance drop, which indicates their effectiveness on regularization for training.

\begin{table*}[!t]
\centering
% \small
\begin{spacing}{1.1}
\resizebox{\textwidth}{30mm}{
\begin{tabular}{lcccccccccc}
\toprule
\multirow{2}{*}{\bf Model} & \multicolumn{5}{c}{\bf GLUE ({\it RoBERTa-large})} & \multicolumn{5}{c}{\bf CLUE ({\it RoBERTa-www-large})} \\
\cmidrule(r){2-6} \cmidrule(r){7-11} 
& {\bf SST-2} & {\bf CoLA} & {\bf MRPC} & {\bf RTE} & {\bf Avg} & {\bf OCNLI} & {\bf IFLYTEK} & {\bf CSL} & {\bf TNEWS} & {\bf Avg}  \\
\midrule
FineTuning & 96.6 & 67.4 & 91.9 & 87.7 & 85.90 & 78.1 & 61.8 & 86.1 & 59.0 & 71.25  \\
\midrule
% R4F \dag & 97.1 & 70.6 & 90.9 & 88.8 & xx & - & - & - & - & - \\
TokenCutoff \dag & 96.9 & {70.0} & 90.9 & 90.6 & 87.10 & 78.2 & 61.8 & 86.1 & {59.2} & 71.33 \\
% S-TokenCutoff & 96.7 & 65.7 & 91.0 & 89.5 & 85.73 & \\
R-drop \dag & 96.9 & {70.0} & 91.4 & 88.4 & 86.67 & 78.9 & 61.6 & 86.6 & 58.9 & 71.50  \\
R3F \dag & {97.0} & 71.2 & 91.6 & 88.5 & 87.07 & - & - & - & - & - \\
PostTraining & 95.0 & 64.7 & 91.2 & 84.1 & 83.75 & 76.5 & 62.1 & 87.0 & 58.9 & 71.13 \\
\midrule
TDT w/o $\mathcal{L}_C$ & 96.4 & 69.3 & 91.9 & 89.5 & 86.77 & 78.6 & 61.9 & 86.9 & 59.0 & 71.60 \\
TDT w/o $\mathcal{L}_R$ & 96.4 & 66.7 & 91.4 & 90.6 & 86.28 & 79.2 & 62.1 & 86.9 & 58.9 & 71.77 \\
TDT-{\it hard} & 96.7 & 67.6 & 92.2 & 90.3 & 86.70 & 79.1 & 62.5 & 87.0 & 59.1 & 71.93 \\
TDT & {97.0} & 69.3 & {93.1} & {91.0} & \textbf{87.60} & {79.5} & {62.9} & {87.2} & {59.2} & \textbf{72.20} \\
\bottomrule
\end{tabular}}
\end{spacing}
\caption{
\label{tab_results_discuss}
Results of RoBERTa-large trained with TDT, variants or previous methods on 4 GLUE tasks and 4 CLUE tasks. 
For GLUE, results with {\dag} are taken from 
% indicate 
% that we directly report the scores from 
the corresponding paper. 
% Other results are from our implementation.
% Other results are from our implementation.
}
\end{table*}

\paragraph{Soft Perturbation vs. Hard Perturbation.}
The confidence score in this paper is continuous value ranging from 0 to 1, and we perturb the input in a soft way.
It is straightforward to investigate the discrete counterpart.
To this end, we model the discrete confidence score with the Gumbel-Softmax trick \cite{jang2017gumbel}.
% , which is an approximation to sampling from the {\it argmax}.
More detailed is introduced in Appendix \ref{sec:appendixD}.
We denote the model trained with the hard strategy as TDT-{\it hard} and show the comparison in Table \ref{tab_results_discuss}. 
From the table, 
% we can see that 
both TDT-{\it hard} and TDT yield better performance than vanilla fine-tuning.
This observation supports our claim that different tokens or phrases contribute differently to the final results, which can be detected by task-guided signal and then used to model more reliable encoders by our proposed regularization.
Moreover, the inferior performance of TDT-{\it hard} shows that naively removing tokens has an adverse effect on context modeling and thus it is better to regularize the over-reliance in a soft manner.

\subsection{Compared with Previous Methods}
\paragraph{TDT vs. Token Cutoff.}
Our method can also be viewed as a soft variant of token cutoff \cite{shen2020tokencutoff}, which is a data augmentation strategy.
Table \ref{tab_results_discuss} shows the results where we find that {\it TDT} performs better than {\it TokenCutoff}, which demonstrates that
the improvement of our method is not entirely due to the effect of data augmentation but stems from the design of the training objectives.

\paragraph{TDT vs. R-drop \& R3F.}

Recently, \citet{liang2021rdrop} proposed R-drop to regularize the consistency of sub-models obtained through dropout.
\citet{aghajanyan2021r3f} introduced R3F rooted in trust region theory, which adds noise into the input embedding and minimize the KL divergence between prediction distributions given original input and noisy input.
Both of them are task-agnostic, while
our proposed method constructs two derived variants with task signal, and concentrates on how to disentangle the task-relevant and task-irrelevant factors.
% find reliable task-specific patterns 
% concentrates on how to enable PLMs to quickly find reliable task-specific patterns for downstream tasks.
The better performance of TDT compared with the strong R-drop and R3F baselines (Table \ref{tab_results_discuss}) verify the advantage of task-driven regularization.

\paragraph{TDT vs. Post-Training.}
Post-training is an effective approach to reduce the objective gap between pretrained model and downstream tasks \cite{gu-etal-2020-train}, which continues to train PLMs on task (or in-domain) training data with mask language model (MLM) loss.
The difference lies in that we focus on the fine-tuning stage.
Here, we compare TDT with the model first post-trained via MLM on training set of each task and then fine-tuned.
It is surprising that post-training does not always have a positive effect on downstream fine-tuning, while TDT shows effective performance without additional post-training time consumption. 

\section{Related Work}
Fine-tuning large-scale PLMs tends to be a popular paradigm of various NLP tasks \cite{devlin-etal-2019-bert,liu-etal-2019-robust,yang2019xlnet}. 
However, 
% recent studies have found that 
the fine-tuned models fail to capture task-specific patterns due to the imbalanced nature between the large number of parameters and limited training data \cite{aghajanyan2020better}.
To address this issue, two main research lines are proposed: (1) continual pretraining after general pre-training, (2) regularization techniques in fine-tuning.

Continual pretraining of PLMs on unlabeled data of a given downstream domain or task has been proved effective for the end-task performance \cite{gururangan-etal-2020-dont}, and various continual pre-training objectives designed for different downstream tasks have been proposed \cite{tian-etal-2020-skep,wu-etal-2021-domain}. 
For example, \citet{gu-etal-2020-train} propose a selective masking strategy to learn task-specific patterns based on mid-scale in-domain data.
However, such methods usually rely on extra in-domain data and manually designed training objectives.

% On the other hand, 
% given that fine-tuned models suffer from overfitting problems, different fine-tuning techniques have been proposed.
Due to the overfitting problems of fine-tuning, lots of regularization techniques have been proposed.
\citet{lee2019mixout} and \citet{chen-etal-2020-recall} regularize fine-tuned weights with original pretrained weights while others design adversarial training objectives or introduce noise into the input \cite{Zhu2019FreeLB,jiang-etal-2020-smart,aghajanyan2020better,shen2020tokencutoff,yu-etal-2021-fine,hua-etal-2021-noise,qu2020coda}.
\citet{liang2021rdrop} regularize the training by minimizing the KL-divergence between the output distributions of two sub-models sampled by dropout and \citet{xu2021raise} only updates a sub-set of the whole network during fine-tuning by selectively masking out the gradients in both task-free and task-driven ways.
\citet{moon2020masker} handle the over-reliance problem by reconstructing keywords based on other words and making low-confidence predictions without enough context.

\section{Conclusion}
In this paper, we propose task-guided disentangled tuning for enhancing the efficiency and robustness of PLMs in downstream NLP tasks. 
Our method is able to efficiently distinguish task-specific features and task-agnostic ones, and bridges the gap between pretraining and adaptation without the need of immediate continual training.
% to promote PLMs to better at modeling task-aware salient context, and hence strengthens the robustness of the entire model.
% a universal task-guided tuning which aims to help pretrained language models find informative inputs for a certain downstream tasks 
% and for the over-reliance problem of fine-tuned models, a constrastive based regularization is further proposed to enhance our method. 
Experiments on GLUE and CLUE benchmarks demonstrate the effectiveness of our method, and extensive analysis shows the advantage in domain generalization and low-resource setting over fine-tuning.

% \section*{Acknowledgements}

\bibliography{anthology,custom}

\begin{thebibliography}{43}
\expandafter\ifx\csname natexlab\endcsname\relax\def\natexlab#1{#1}\fi

\bibitem[{Aghajanyan et~al.(2021{\natexlab{a}})Aghajanyan, Gupta, Shrivastava,
  Chen, Zettlemoyer, and Gupta}]{aghajanyan2021muppet}
Armen Aghajanyan, Anchit Gupta, Akshat Shrivastava, Xilun Chen, Luke
  Zettlemoyer, and Sonal Gupta. 2021{\natexlab{a}}.
\newblock Muppet: Massive multi-task representations with pre-finetuning.
\newblock \emph{arXiv preprint arXiv:2101.11038}.

\bibitem[{Aghajanyan et~al.(2020)Aghajanyan, Shrivastava, Gupta, Goyal,
  Zettlemoyer, and Gupta}]{aghajanyan2020better}
Armen Aghajanyan, Akshat Shrivastava, Anchit Gupta, Naman Goyal, Luke
  Zettlemoyer, and Sonal Gupta. 2020.
\newblock Better fine-tuning by reducing representational collapse.
\newblock \emph{arXiv preprint arXiv:2008.03156}.

\bibitem[{Aghajanyan et~al.(2021{\natexlab{b}})Aghajanyan, Shrivastava, Gupta,
  Goyal, Zettlemoyer, and Gupta}]{aghajanyan2021r3f}
Armen Aghajanyan, Akshat Shrivastava, Anchit Gupta, Naman Goyal, Luke
  Zettlemoyer, and Sonal Gupta. 2021{\natexlab{b}}.
\newblock \href {https://openreview.net/forum?id=OQ08SN70M1V} {Better
  fine-tuning by reducing representational collapse}.
\newblock In \emph{9th International Conference on Learning Representations,
  {ICLR} 2021, Virtual Event, Austria, May 3-7, 2021}. OpenReview.net.

\bibitem[{Chen et~al.(2020)Chen, Hou, Cui, Che, Liu, and
  Yu}]{chen-etal-2020-recall}
Sanyuan Chen, Yutai Hou, Yiming Cui, Wanxiang Che, Ting Liu, and Xiangzhan Yu.
  2020.
\newblock \href {https://doi.org/10.18653/v1/2020.emnlp-main.634} {Recall and
  learn: Fine-tuning deep pretrained language models with less forgetting}.
\newblock In \emph{Proceedings of the 2020 Conference on Empirical Methods in
  Natural Language Processing (EMNLP)}, pages 7870--7881, Online. Association
  for Computational Linguistics.

\bibitem[{Clark et~al.(2020)Clark, Luong, Le, and Manning}]{clark2020electra}
Kevin Clark, Minh-Thang Luong, Quoc~V. Le, and Christopher~D. Manning. 2020.
\newblock \href {https://openreview.net/pdf?id=r1xMH1BtvB} {{ELECTRA}:
  Pre-training text encoders as discriminators rather than generators}.
\newblock In \emph{ICLR}.

\bibitem[{Cui et~al.(2020)Cui, Che, Liu, Qin, Wang, and
  Hu}]{cui-etal-2020-revisiting}
Yiming Cui, Wanxiang Che, Ting Liu, Bing Qin, Shijin Wang, and Guoping Hu.
  2020.
\newblock \href {https://doi.org/10.18653/v1/2020.findings-emnlp.58}
  {Revisiting pre-trained models for {C}hinese natural language processing}.
\newblock In \emph{Findings of the Association for Computational Linguistics:
  EMNLP 2020}, pages 657--668, Online. Association for Computational
  Linguistics.

\bibitem[{Cui et~al.(2019)Cui, Che, Liu, Qin, Yang, Wang, and Hu}]{cui2019pre}
Yiming Cui, Wanxiang Che, Ting Liu, Bing Qin, Ziqing Yang, Shijin Wang, and
  Guoping Hu. 2019.
\newblock Pre-training with whole word masking for chinese bert.
\newblock \emph{arXiv preprint arXiv:1906.08101}.

\bibitem[{Devlin et~al.(2019)Devlin, Chang, Lee, and
  Toutanova}]{devlin-etal-2019-bert}
Jacob Devlin, Ming-Wei Chang, Kenton Lee, and Kristina Toutanova. 2019.
\newblock \href {https://doi.org/10.18653/v1/N19-1423} {{BERT}: Pre-training of
  deep bidirectional transformers for language understanding}.
\newblock In \emph{Proceedings of the 2019 Conference of the North {A}merican
  Chapter of the Association for Computational Linguistics: Human Language
  Technologies, Volume 1 (Long and Short Papers)}, pages 4171--4186,
  Minneapolis, Minnesota. Association for Computational Linguistics.

\bibitem[{Fong and Vedaldi(2017)}]{Fong2017interpretable}
Ruth~C. Fong and Andrea Vedaldi. 2017.
\newblock \href {https://doi.org/10.1109/ICCV.2017.371} {Interpretable
  explanations of black boxes by meaningful perturbation}.
\newblock In \emph{{IEEE} International Conference on Computer Vision, {ICCV}
  2017, Venice, Italy, October 22-29, 2017}, pages 3449--3457. {IEEE} Computer
  Society.

\bibitem[{Geirhos et~al.(2020)Geirhos, Jacobsen, Michaelis, Zemel, Brendel,
  Bethge, and Wichmann}]{geirhos2020shortcut}
Robert Geirhos, J{\"{o}}rn{-}Henrik Jacobsen, Claudio Michaelis, Richard~S.
  Zemel, Wieland Brendel, Matthias Bethge, and Felix~A. Wichmann. 2020.
\newblock \href {http://arxiv.org/abs/2004.07780} {Shortcut learning in deep
  neural networks}.
\newblock \emph{CoRR}, abs/2004.07780.

\bibitem[{Gu et~al.(2020)Gu, Zhang, Wang, Liu, and Sun}]{gu-etal-2020-train}
Yuxian Gu, Zhengyan Zhang, Xiaozhi Wang, Zhiyuan Liu, and Maosong Sun. 2020.
\newblock \href {https://doi.org/10.18653/v1/2020.emnlp-main.566} {Train no
  evil: Selective masking for task-guided pre-training}.
\newblock In \emph{Proceedings of the 2020 Conference on Empirical Methods in
  Natural Language Processing (EMNLP)}, pages 6966--6974, Online. Association
  for Computational Linguistics.

\bibitem[{Gururangan et~al.(2020)Gururangan, Marasovi{\'c}, Swayamdipta, Lo,
  Beltagy, Downey, and Smith}]{gururangan-etal-2020-dont}
Suchin Gururangan, Ana Marasovi{\'c}, Swabha Swayamdipta, Kyle Lo, Iz~Beltagy,
  Doug Downey, and Noah~A. Smith. 2020.
\newblock \href {https://doi.org/10.18653/v1/2020.acl-main.740} {Don{'}t stop
  pretraining: Adapt language models to domains and tasks}.
\newblock In \emph{Proceedings of the 58th Annual Meeting of the Association
  for Computational Linguistics}, pages 8342--8360, Online. Association for
  Computational Linguistics.

\bibitem[{He et~al.(2020)He, Liu, Gao, and Chen}]{he2020deberta}
Pengcheng He, Xiaodong Liu, Jianfeng Gao, and Weizhu Chen. 2020.
\newblock Deberta: Decoding-enhanced bert with disentangled attention.
\newblock \emph{arXiv preprint arXiv:2006.03654}.

\bibitem[{Hua et~al.(2021)Hua, Li, Dou, Xu, and Luo}]{hua-etal-2021-noise}
Hang Hua, Xingjian Li, Dejing Dou, Chengzhong Xu, and Jiebo Luo. 2021.
\newblock \href {https://doi.org/10.18653/v1/2021.naacl-main.258} {Noise
  stability regularization for improving {BERT} fine-tuning}.
\newblock In \emph{Proceedings of the 2021 Conference of the North American
  Chapter of the Association for Computational Linguistics: Human Language
  Technologies}, pages 3229--3241, Online. Association for Computational
  Linguistics.

\bibitem[{Jang et~al.(2017)Jang, Gu, and Poole}]{jang2017gumbel}
Eric Jang, Shixiang Gu, and Ben Poole. 2017.
\newblock \href {https://openreview.net/forum?id=rkE3y85ee} {Categorical
  reparameterization with gumbel-softmax}.
\newblock In \emph{5th International Conference on Learning Representations,
  {ICLR} 2017, Toulon, France, April 24-26, 2017, Conference Track
  Proceedings}. OpenReview.net.

\bibitem[{Jiang et~al.(2020)Jiang, He, Chen, Liu, Gao, and
  Zhao}]{jiang-etal-2020-smart}
Haoming Jiang, Pengcheng He, Weizhu Chen, Xiaodong Liu, Jianfeng Gao, and Tuo
  Zhao. 2020.
\newblock \href {https://doi.org/10.18653/v1/2020.acl-main.197} {{SMART}:
  Robust and efficient fine-tuning for pre-trained natural language models
  through principled regularized optimization}.
\newblock In \emph{Proceedings of the 58th Annual Meeting of the Association
  for Computational Linguistics}, pages 2177--2190, Online. Association for
  Computational Linguistics.

\bibitem[{Kang et~al.(2020)Kang, Han, and Hwang}]{kang-etal-2020-neural}
Minki Kang, Moonsu Han, and Sung~Ju Hwang. 2020.
\newblock \href {https://doi.org/10.18653/v1/2020.emnlp-main.493} {Neural mask
  generator: Learning to generate adaptive word maskings for language model
  adaptation}.
\newblock In \emph{Proceedings of the 2020 Conference on Empirical Methods in
  Natural Language Processing (EMNLP)}, pages 6102--6120, Online. Association
  for Computational Linguistics.

\bibitem[{Lan et~al.(2019)Lan, Chen, Goodman, Gimpel, Sharma, and
  Soricut}]{lan2019albert}
Zhenzhong Lan, Mingda Chen, Sebastian Goodman, Kevin Gimpel, Piyush Sharma, and
  Radu Soricut. 2019.
\newblock Albert: A lite bert for self-supervised learning of language
  representations.
\newblock \emph{arXiv preprint arXiv:1909.11942}.

\bibitem[{Lee et~al.(2019)Lee, Cho, and Kang}]{lee2019mixout}
Cheolhyoung Lee, Kyunghyun Cho, and Wanmo Kang. 2019.
\newblock Mixout: Effective regularization to finetune large-scale pretrained
  language models.
\newblock \emph{arXiv preprint arXiv:1909.11299}.

\bibitem[{Lee et~al.(2020)Lee, Cho, and Kang}]{Lee2020Mixout}
Cheolhyoung Lee, Kyunghyun Cho, and Wanmo Kang. 2020.
\newblock \href {https://openreview.net/forum?id=HkgaETNtDB} {Mixout: Effective
  regularization to finetune large-scale pretrained language models}.
\newblock In \emph{8th International Conference on Learning Representations,
  {ICLR} 2020, Addis Ababa, Ethiopia, April 26-30, 2020}. OpenReview.net.

\bibitem[{Liang et~al.(2021)Liang, Wu, Li, Wang, Meng, Qin, Chen, Zhang, and
  Liu}]{liang2021rdrop}
Xiaobo Liang, Lijun Wu, Juntao Li, Yue Wang, Qi~Meng, Tao Qin, Wei Chen, Min
  Zhang, and Tie{-}Yan Liu. 2021.
\newblock \href {http://arxiv.org/abs/2106.14448} {R-drop: Regularized dropout
  for neural networks}.
\newblock \emph{CoRR}, abs/2106.14448.

\bibitem[{Liu et~al.(2019{\natexlab{a}})Liu, Ma, Huang, Xiong, and
  He}]{liu-etal-2019-robust}
Hairong Liu, Mingbo Ma, Liang Huang, Hao Xiong, and Zhongjun He.
  2019{\natexlab{a}}.
\newblock \href {https://doi.org/10.18653/v1/P19-1291} {Robust neural machine
  translation with joint textual and phonetic embedding}.
\newblock In \emph{Proceedings of the 57th Annual Meeting of the Association
  for Computational Linguistics}, pages 3044--3049, Florence, Italy.
  Association for Computational Linguistics.

\bibitem[{Liu et~al.(2019{\natexlab{b}})Liu, Ott, Goyal, Du, Joshi, Chen, Levy,
  Lewis, Zettlemoyer, and Stoyanov}]{liu2019roberta}
Yinhan Liu, Myle Ott, Naman Goyal, Jingfei Du, Mandar Joshi, Danqi Chen, Omer
  Levy, Mike Lewis, Luke Zettlemoyer, and Veselin Stoyanov. 2019{\natexlab{b}}.
\newblock Roberta: A robustly optimized bert pretraining approach.
\newblock \emph{arXiv preprint arXiv:1907.11692}.

\bibitem[{McCoy et~al.(2019)McCoy, Pavlick, and Linzen}]{mccoy-etal-2019-right}
Tom McCoy, Ellie Pavlick, and Tal Linzen. 2019.
\newblock \href {https://doi.org/10.18653/v1/P19-1334} {Right for the wrong
  reasons: Diagnosing syntactic heuristics in natural language inference}.
\newblock In \emph{Proceedings of the 57th Annual Meeting of the Association
  for Computational Linguistics}, pages 3428--3448, Florence, Italy.
  Association for Computational Linguistics.

\bibitem[{Min et~al.(2019)Min, Wallace, Singh, Gardner, Hajishirzi, and
  Zettlemoyer}]{min-etal-2019-compositional}
Sewon Min, Eric Wallace, Sameer Singh, Matt Gardner, Hannaneh Hajishirzi, and
  Luke Zettlemoyer. 2019.
\newblock \href {https://doi.org/10.18653/v1/P19-1416} {Compositional questions
  do not necessitate multi-hop reasoning}.
\newblock In \emph{Proceedings of the 57th Annual Meeting of the Association
  for Computational Linguistics}, pages 4249--4257, Florence, Italy.
  Association for Computational Linguistics.

\bibitem[{Moon et~al.(2020)Moon, Mo, Lee, Lee, and Shin}]{moon2020masker}
Seung~Jun Moon, Sangwoo Mo, Kimin Lee, Jaeho Lee, and Jinwoo Shin. 2020.
\newblock Masker: Masked keyword regularization for reliable text
  classification.
\newblock \emph{arXiv preprint arXiv:2012.09392}.

\bibitem[{Niven and Kao(2019)}]{niven-kao-2019-probing}
Timothy Niven and Hung-Yu Kao. 2019.
\newblock \href {https://doi.org/10.18653/v1/P19-1459} {Probing neural network
  comprehension of natural language arguments}.
\newblock In \emph{Proceedings of the 57th Annual Meeting of the Association
  for Computational Linguistics}, pages 4658--4664, Florence, Italy.
  Association for Computational Linguistics.

\bibitem[{Qu et~al.(2020)Qu, Shen, Shen, Sajeev, Han, and Chen}]{qu2020coda}
Yanru Qu, Dinghan Shen, Yelong Shen, Sandra Sajeev, Jiawei Han, and Weizhu
  Chen. 2020.
\newblock Coda: Contrast-enhanced and diversity-promoting data augmentation for
  natural language understanding.
\newblock \emph{arXiv preprint arXiv:2010.08670}.

\bibitem[{Shen et~al.(2020)Shen, Zheng, Shen, Qu, and
  Chen}]{shen2020tokencutoff}
Dinghan Shen, Mingzhi Zheng, Yelong Shen, Yanru Qu, and Weizhu Chen. 2020.
\newblock \href {http://arxiv.org/abs/2009.13818} {A simple but tough-to-beat
  data augmentation approach for natural language understanding and
  generation}.
\newblock \emph{CoRR}, abs/2009.13818.

\bibitem[{Sun et~al.(2019)Sun, Qiu, Xu, and Huang}]{sun2019shortcutnlp}
Chi Sun, Xipeng Qiu, Yige Xu, and Xuanjing Huang. 2019.
\newblock \href {https://doi.org/10.1007/978-3-030-32381-3\_16} {How to
  fine-tune {BERT} for text classification?}
\newblock In \emph{Chinese Computational Linguistics - 18th China National
  Conference, {CCL} 2019, Kunming, China, October 18-20, 2019, Proceedings},
  volume 11856 of \emph{Lecture Notes in Computer Science}, pages 194--206.
  Springer.

\bibitem[{Tian et~al.(2020)Tian, Gao, Xiao, Liu, He, Wu, Wang, and
  Wu}]{tian-etal-2020-skep}
Hao Tian, Can Gao, Xinyan Xiao, Hao Liu, Bolei He, Hua Wu, Haifeng Wang, and
  Feng Wu. 2020.
\newblock \href {https://doi.org/10.18653/v1/2020.acl-main.374} {{SKEP}:
  Sentiment knowledge enhanced pre-training for sentiment analysis}.
\newblock In \emph{Proceedings of the 58th Annual Meeting of the Association
  for Computational Linguistics}, pages 4067--4076, Online. Association for
  Computational Linguistics.

\bibitem[{van~der Maaten and Hinton(2008)}]{maaten2008sne}
Laurens van~der Maaten and Geoffrey Hinton. 2008.
\newblock \href {http://jmlr.org/papers/v9/vandermaaten08a.html} {Visualizing
  data using t-sne}.
\newblock \emph{Journal of Machine Learning Research}, 9(86):2579--2605.

\bibitem[{Voita et~al.(2019)Voita, Talbot, Moiseev, Sennrich, and
  Titov}]{voita-etal-2019-analyzing}
Elena Voita, David Talbot, Fedor Moiseev, Rico Sennrich, and Ivan Titov. 2019.
\newblock \href {https://doi.org/10.18653/v1/P19-1580} {Analyzing multi-head
  self-attention: Specialized heads do the heavy lifting, the rest can be
  pruned}.
\newblock In \emph{Proceedings of the 57th Annual Meeting of the Association
  for Computational Linguistics}, pages 5797--5808, Florence, Italy.
  Association for Computational Linguistics.

\bibitem[{Wang et~al.(2019)Wang, Hula, Xia, Pappagari, McCoy, Patel, Kim,
  Tenney, Huang, Yu, Jin, Chen, Van~Durme, Grave, Pavlick, and
  Bowman}]{wang-etal-2019-tell}
Alex Wang, Jan Hula, Patrick Xia, Raghavendra Pappagari, R.~Thomas McCoy, Roma
  Patel, Najoung Kim, Ian Tenney, Yinghui Huang, Katherin Yu, Shuning Jin,
  Berlin Chen, Benjamin Van~Durme, Edouard Grave, Ellie Pavlick, and Samuel~R.
  Bowman. 2019.
\newblock \href {https://doi.org/10.18653/v1/P19-1439} {Can you tell me how to
  get past sesame street? sentence-level pretraining beyond language modeling}.
\newblock In \emph{Proceedings of the 57th Annual Meeting of the Association
  for Computational Linguistics}, pages 4465--4476, Florence, Italy.
  Association for Computational Linguistics.

\bibitem[{Wang et~al.(2018)Wang, Singh, Michael, Hill, Levy, and
  Bowman}]{wang-etal-2018-glue}
Alex Wang, Amanpreet Singh, Julian Michael, Felix Hill, Omer Levy, and Samuel
  Bowman. 2018.
\newblock \href {https://doi.org/10.18653/v1/W18-5446} {{GLUE}: A multi-task
  benchmark and analysis platform for natural language understanding}.
\newblock In \emph{Proceedings of the 2018 {EMNLP} Workshop {B}lackbox{NLP}:
  Analyzing and Interpreting Neural Networks for {NLP}}, pages 353--355,
  Brussels, Belgium. Association for Computational Linguistics.

\bibitem[{Wu et~al.(2021)Wu, Xu, Song, Jin, Zhang, and
  Song}]{wu-etal-2021-domain}
Han Wu, Kun Xu, Linfeng Song, Lifeng Jin, Haisong Zhang, and Linqi Song. 2021.
\newblock \href {https://doi.org/10.18653/v1/2021.acl-short.84}
  {Domain-adaptive pretraining methods for dialogue understanding}.
\newblock In \emph{Proceedings of the 59th Annual Meeting of the Association
  for Computational Linguistics and the 11th International Joint Conference on
  Natural Language Processing (Volume 2: Short Papers)}, pages 665--669,
  Online. Association for Computational Linguistics.

\bibitem[{Xu et~al.(2020)Xu, Hu, Zhang, Li, Cao, Li, Xu, Sun, Yu, Yu, Tian,
  Dong, Liu, Shi, Cui, Li, Zeng, Wang, Xie, Li, Patterson, Tian, Zhang, Zhou,
  Liu, Zhao, Zhao, Yue, Zhang, Yang, Richardson, and Lan}]{xu2020CLUE}
Liang Xu, Hai Hu, Xuanwei Zhang, Lu~Li, Chenjie Cao, Yudong Li, Yechen Xu, Kai
  Sun, Dian Yu, Cong Yu, Yin Tian, Qianqian Dong, Weitang Liu, Bo~Shi, Yiming
  Cui, Junyi Li, Jun Zeng, Rongzhao Wang, Weijian Xie, Yanting Li, Yina
  Patterson, Zuoyu Tian, Yiwen Zhang, He~Zhou, Shaoweihua Liu, Zhe Zhao, Qipeng
  Zhao, Cong Yue, Xinrui Zhang, Zhengliang Yang, Kyle Richardson, and Zhenzhong
  Lan. 2020.
\newblock \href {https://doi.org/10.18653/v1/2020.coling-main.419} {{CLUE:} {A}
  chinese language understanding evaluation benchmark}.
\newblock In \emph{Proceedings of the 28th International Conference on
  Computational Linguistics, {COLING} 2020, Barcelona, Spain (Online), December
  8-13, 2020}, pages 4762--4772. International Committee on Computational
  Linguistics.

\bibitem[{Xu et~al.(2021{\natexlab{a}})Xu, Lu, Yuan, Zhang, Yuan, Xu, Wei, Pan,
  and Hu}]{Xu2021Fewclue}
Liang Xu, Xiaojing Lu, Chenyang Yuan, Xuanwei Zhang, Hu~Yuan, Huilin Xu, Guoao
  Wei, Xiang Pan, and Hai Hu. 2021{\natexlab{a}}.
\newblock \href {http://arxiv.org/abs/2107.07498} {Fewclue: {A} chinese
  few-shot learning evaluation benchmark}.
\newblock \emph{CoRR}, abs/2107.07498.

\bibitem[{Xu et~al.(2021{\natexlab{b}})Xu, Luo, Zhang, Tan, Chang, Huang, and
  Huang}]{xu2021raise}
Runxin Xu, Fuli Luo, Zhiyuan Zhang, Chuanqi Tan, Baobao Chang, Songfang Huang,
  and Fei Huang. 2021{\natexlab{b}}.
\newblock Raise a child in large language model: Towards effective and
  generalizable fine-tuning.
\newblock \emph{arXiv preprint arXiv:2109.05687}.

\bibitem[{Yang et~al.(2019)Yang, Dai, Yang, Carbonell, Salakhutdinov, and
  Le}]{yang2019xlnet}
Zhilin Yang, Zihang Dai, Yiming Yang, Jaime Carbonell, Russ~R Salakhutdinov,
  and Quoc~V Le. 2019.
\newblock Xlnet: Generalized autoregressive pretraining for language
  understanding.
\newblock \emph{Advances in neural information processing systems}, 32.

\bibitem[{Ye et~al.(2021)Ye, Li, Wang, Bolte, Ma, Yih, Ren, and
  Khabsa}]{ye2021influence}
Qinyuan Ye, Belinda~Z Li, Sinong Wang, Benjamin Bolte, Hao Ma, Wen-tau Yih,
  Xiang Ren, and Madian Khabsa. 2021.
\newblock On the influence of masking policies in intermediate pre-training.
\newblock \emph{arXiv preprint arXiv:2104.08840}.

\bibitem[{Yu et~al.(2021)Yu, Zuo, Jiang, Ren, Zhao, and
  Zhang}]{yu-etal-2021-fine}
Yue Yu, Simiao Zuo, Haoming Jiang, Wendi Ren, Tuo Zhao, and Chao Zhang. 2021.
\newblock \href {https://doi.org/10.18653/v1/2021.naacl-main.84} {Fine-tuning
  pre-trained language model with weak supervision: A contrastive-regularized
  self-training approach}.
\newblock In \emph{Proceedings of the 2021 Conference of the North American
  Chapter of the Association for Computational Linguistics: Human Language
  Technologies}, pages 1063--1077, Online. Association for Computational
  Linguistics.

\bibitem[{Zhu et~al.(2020)Zhu, Cheng, Gan, Sun, Goldstein, and
  Liu}]{Zhu2019FreeLB}
Chen Zhu, Yu~Cheng, Zhe Gan, Siqi Sun, Tom Goldstein, and Jingjing Liu. 2020.
\newblock \href {https://openreview.net/forum?id=BygzbyHFvB} {Freelb: Enhanced
  adversarial training for natural language understanding}.
\newblock In \emph{8th International Conference on Learning Representations,
  {ICLR} 2020, Addis Ababa, Ethiopia, April 26-30, 2020}. OpenReview.net.

\end{thebibliography}
\bibliographystyle{acl_natbib}

\appendix

\section{GLUE and CLUE Benchmark}
\label{sec:appendixA}

\begin{table}[!h]
\centering
\small
\begin{spacing}{1.0}
\begin{tabular}{lccc}
\toprule
{\bf Dataset} & {\bf \# Train} & {\bf \# Dev} & {\bf Metrics} \\
\midrule
\multicolumn{4}{c}{\bf GLUE} \\
MNLI & 393k & 9.8k &Accuracy \\
QQP & 364k & 40k & Accuracy \\
QNLI & 105k & 5.5k & Accuracy \\
SST-2 & 67k & 872 & Accuracy \\
CoLA & 8.5k & 1.0k & Matthews Corr \\
STS-B & 5.7k & 1.5k & Spearman Corr \\
MRPC & 3.7k & 408 & Accuracy \\
RTE & 2.5k & 277 & Accuracy \\
\midrule
\multicolumn{4}{c}{\bf CLUE} \\
OCNLI & 50k & 3k & Accuracy \\
IFLYTEK & 12.1k & 2.6k & Accuracy \\
CSL & 20k & 3k & Accuracy \\
TNEWS & 53.3k & 10k & Accuracy \\
AFQMC & 34.3k & 4.3k & Accuracy \\
CMNLI & 391k & 12k & Accuracy \\
CLUEWSC & 1.2k & 304 & Accuracy \\
\bottomrule
\end{tabular}
\end{spacing}
\caption{
\label{tab_results_data_statistic}
Data Statistics and Evaluate Metrics.
}
\end{table}

In this paper, we conduct experiments on 8 datasets in GLUE benchmark \cite{wang-etal-2018-glue}, and 5 datasets in CLUE \cite{xu2020CLUE}, including the short text classification task TNEWS, the long text classification tasks IFLYTEK and CSL, and sentence-pair classification tasks AFQMC and OCNLI.
The data statistics and evaluate metrics are illustrated in Table \ref{tab_results_data_statistic}.

\section{Settings for Different Pretrained Models} \label{sec:appendixB}

\begin{table}[!h]
\centering
\small
\begin{spacing}{1.0}
\begin{tabular}{lcccc}
\toprule
{\bf Task} & {\bf Batch Size} & {\bf Steps} & {\bf Warmup} & {\bf lr} \\
\midrule
\multicolumn{5}{c}{\bf GLUE} \\
\multicolumn{5}{c}{\it BERT-base} \\
MNLI & 128 & 10000 & 1000 & 4e-5 \\
QQP & 128 & 10000 & 1000 & 4e-5 \\
QNLI & 64 & 3000 & 300 & 4e-5  \\
SST-2 & 64 & 3000 & 300 & 4e-5 \\
CoLA & 64 & 2000 & 200 & 2e-5 \\
STS-B & 64 & 3000 & 300 & 4e-5 \\
MRPC & 64 & 2000 & 200 & 1e-5 \\
RTE & 64 & 2000 & 200 & 2e-5 \\
\multicolumn{5}{c}{\it BERT-large \& RoBERT-large} \\
MNLI & 64 & 10000 & 1000 & 2e-5 \\
QQP & 64 & 10000 & 1000 & 2e-5 \\
QNLI & 64 & 3000 & 300 & 2e-5 \\
SST-2 & 64 & 3000 & 300 & 2e-5 \\
CoLA & 32 & 3000 & 300 & 2e-5 \\
STS-B & 64 & 3000 & 300 & 2e-5 \\
MRPC & 64 & 2000 & 200 & 2e-5 \\
RTE & 64 & 2000 & 100 & 2e-5 \\
\midrule
\multicolumn{5}{c}{\bf CLUE} \\
\multicolumn{5}{c}{\it BERT-wwm-base} \\
OCNLI & 64 & 3000 & 300 & 4e-5 \\
IFLYTEK & 16 & 5000 & 300 & 3e-5 \\
CSL & 32 & 3000 & 300 & 3e-5 \\
TNEWS & 64 & 5000 & 300 & 3e-5 \\
AFQMC & 32 & 3000 & 300 & 3e-5 \\
\multicolumn{5}{c}{\it MacBERT-large \& RoBERT-wwm-large} \\
OCNLI & 32 & 3000 & 300 & 1e-5 \\
IFLYTEK & 16 & 5000 & 300 & 1e-5 \\
CSL & 32 & 3000 & 300 & 1e-5 \\
TNEWS & 64 & 5000 & 300 & 1e-5 \\
AFQMC & 32 & 3000 & 300 & 1e-5 \\
\bottomrule
\end{tabular}
\end{spacing}
\caption{
\label{tab_settings}
Hyperparameters settings for different pretrained models on variant tasks.
}
\end{table}

In this paper, we fine-tuned different pretrained models with TDT, including BERT-base, BERT-large, RoBERTa-large for GLUE and BERT-wwm-base, MacBERT-large, RoBERTa-wwm-large for CLUE.
The batch size, training steps, warmup steps, and learning rate are listed in Table \ref{tab_settings}.

\section{Label Mapping in Domain Generalization}
\label{sec:appendixC}

QQP has two labels, \textit{duplicate} and \textit{not duplicate}. We map \textit{entailment} to \textit{duplicate} and map both \textit{neutral} and \textit{contradiction} to \textit{not duplicate}.
BUSTM \footnote{https://github.com/xiaobu-coai/BUSTM} is a short text matching task of FewCLUE \cite{Xu2021Fewclue}.
We use the public test set.
BUSTM has two labels, \textit{0} and \textit{1}.
We map \textit{entailment} to label \textit{1}, and map both \textit{neutral} and \textit{contradiction} to label \textit{0}.

\section{Detailed of TDT-{\it hard}}
\label{sec:appendixD}
Gumbel-Softmax trick \cite{jang2017gumbel} is an approximation to sampling from the {\it argmax}.
Formally, we replace Eq. \ref{eq_confidence} by:
\begin{align}
     c_i &= {\rm argmax}(\sigma_{\rm Gumbel}(z(e_i))), \\
    \sigma_{\rm Gumbel}(z_i) &= \frac{{\rm exp}((log(z_i)+g_i)/\tau)}{\sum^{K}_{j=1}{\rm exp}((log(z_j)+g_j)/\tau)},
\end{align}
where $g_i \sim $ Gumbel(0,1), $z(\cdot)$ returns the logits produced for a given input, and $\tau$ is the temperature.
By this way, if $c_i$ is 0, the embedding of the $i$-th token is set to the embedding of the ``[MASK]'' token, otherwise the embedding remains unchanged.

\end{document}